\documentclass[10pt,twocolumn,letterpaper]{article}

\usepackage{cvpr}              
\usepackage{amsmath}
\usepackage{amssymb}
\usepackage[accsupp]{axessibility}
\usepackage{enumitem}
\usepackage{algorithm}
\usepackage{algorithmic}
\usepackage{multirow}
\usepackage{booktabs}
\usepackage{tabularx}
\usepackage{makecell}
\usepackage{xcolor}

\definecolor{cvprblue}{rgb}{0.21,0.49,0.74}
\usepackage[pagebackref,breaklinks,colorlinks,allcolors=cvprblue]{hyperref}

\def\paperID{38670} 
\def\confName{CVPR}
\def\confYear{2026}

\title{Breaking the Illusion: When Positive Meets Negative in Multimodal Decoding}

\author{
Yubo Jiang$^{1,2}$ \quad
Yitong An$^{1}$ \quad
Xin Yang$^{2}$ \quad
Abudukelimu Wuerkaixi$^{2}$ \quad
Xuxin Cheng$^{2}$ \quad
Fengying Xie$^{1,3}$ \\
Zhiguo Jiang$^{3}$ \quad
Cao Liu$^{2}$ \quad
Ke Zeng$^{2~\dagger}$ \quad
Haopeng Zhang$^{1,3~\dagger}$\\[0.5em]
$^{1}$School of Astronautics, Beihang University, Beijing 102206, China\\
$^{2}$Longcat Interaction Team, Meituan, Beijing 100102, China \\
$^{3}$Tianmushan Laboratory, Beihang University, Hangzhou 311115, China\\[0.5em]
{\tt\small \{jbond0409, zhanghaopeng\}@buaa.edu.cn (Y.J., H.Z.)}
}

\begin{document}
\maketitle
\begin{abstract}
Vision-Language Models (VLMs) are frequently undermined by object hallucination—generating content that contradicts visual reality—due to an over-reliance on linguistic priors. We introduce Positive-and-Negative Decoding (PND), a training-free inference framework that intervenes directly in the decoding process to enforce visual fidelity. PND is motivated by our key finding of a critical attention deficit in VLMs, where visual features are empirically under-weighted. Our framework corrects this via a dual-path contrast: The positive path amplifies salient visual evidence using multi-layer attention to encourage faithful descriptions, directly counteracting the attention deficit. Simultaneously, the negative path identifies and degrades the core object's features to create a strong counterfactual, which penalizes ungrounded, prior-dominant generation. By contrasting the model's outputs from these two perspectives at each step, PND steers generation towards text that is not just linguistically probable, but visually factual. Extensive experiments on benchmarks like POPE, MME, and CHAIR show that PND achieves state-of-the-art performance with up to 6.5\% accuracy improvement, substantially reducing object hallucination while also enhancing descriptive detail—all without requiring any model retraining. The method generalizes effectively across diverse VLM architectures including LLaVA, InstructBLIP, InternVL, and Qwen-VL. Project Page: 
https://github.com/JiangYubo4399/PND. 

\end{abstract}


\begingroup
\renewcommand\thefootnote{}
\footnotetext{$\dagger$ indicates the corresponding authors.}
\endgroup
\section{Introduction}

\begin{figure}[t]
\centering

\includegraphics[width=0.8\columnwidth]{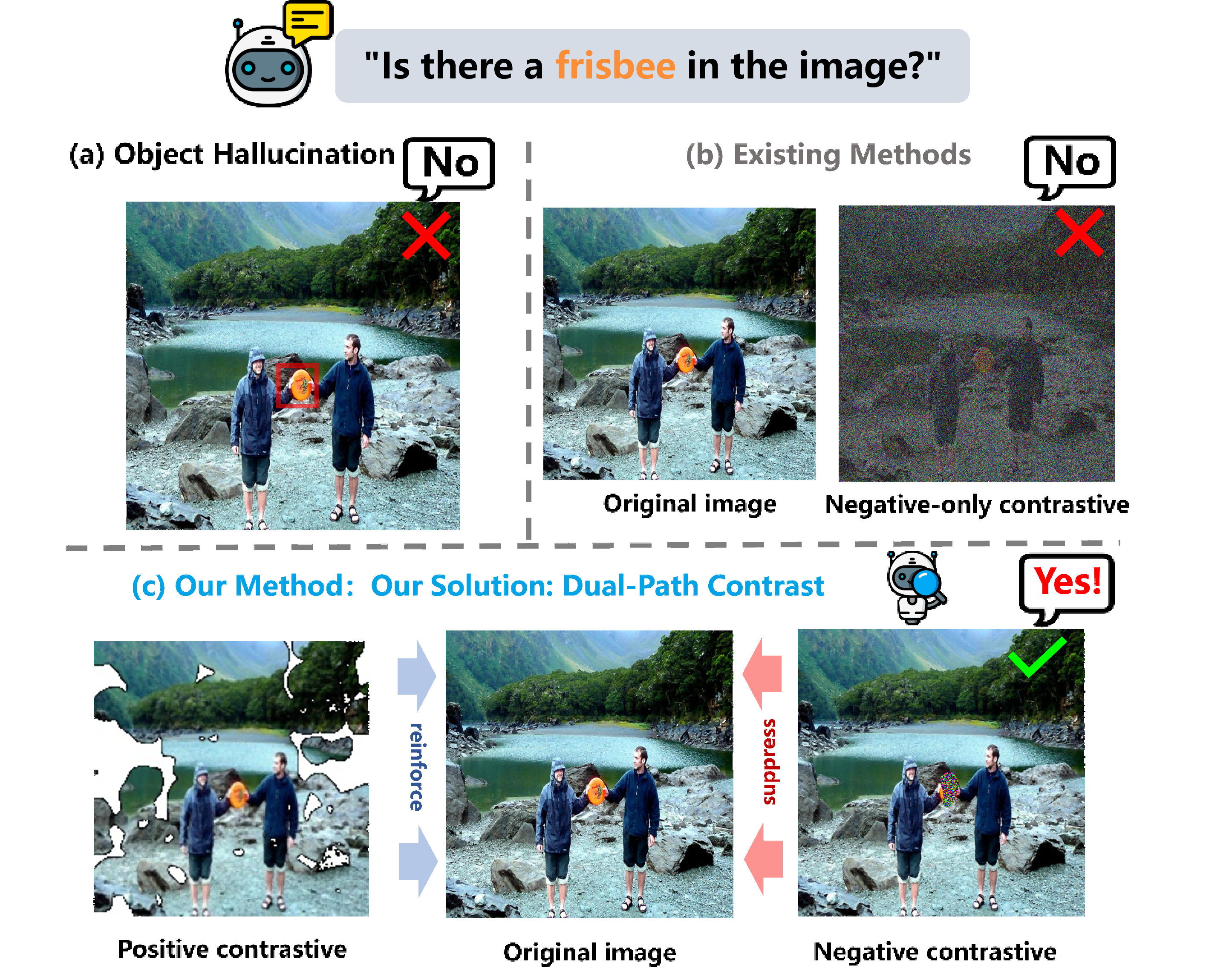} 
\caption{PND suppresses object hallucination via \textbf{dual-path contrast}. (a) A standard VLM fails to identify the frisbee due to weak linguistic priors. (b) Existing negative-only methods are insufficient. (c) Our PND's dual pathways overcome this: the positive path reinforces the object's presence, while the negative path creates a counterfactual, leading to correct identification.}
\label{fig:intro}
\end{figure}

Large-scale vision language models (VLMs) have achieved remarkable success in multimodal tasks \cite{zhou2022learning, zhang2024vision, dai2023instructblip}. However, a critical failure mode persists: these models frequently hallucinate, generating plausible but factually incorrect content that contradicts the visual input \cite{huang2025survey, li2023evaluating, gunjal2024detecting}. We argue this failure is, in essence, a Bayesian reasoning imbalance \cite{watanabe2024recent, cai2008comment}. From a Bayesian perspective, a VLM's generative process is determined by two competing forces: a language prior, referring to the model’s learned co-occurrence biases from 
pre-training that encode how words and visual concepts are statistically aligned; and a visual likelihood, referring to the image-grounded evidence that directly 
constrains what objects and attributes are actually present.
 Hallucination occurs when this balance fails and the generation becomes "prior-dominant" \cite{shi2023towards, liang2022mind}. This imbalance manifests in two primary ways:
\begin{itemize}[leftmargin=*,itemsep=1pt,topsep=2pt]

    \item \textbf{Object Confabulation (Positive Hallucination).}
    Positive hallucination refers to the model \emph{inventing} objects that are not present,
    typically when a dominant language prior overrules visual evidence. Existing
    perturbation-only approaches, such as VCD-style negative-pair construction, attempt
    to suppress this behavior but often degrade the image too heavily, removing crucial
    semantics needed for reliable grounding.

    \item \textbf{Object Omission (Negative Hallucination).}
    In contrast, omission occurs when a real object receives insufficient grounding.
    As shown in \cref{fig:intro}(b), the frisbee is clearly visible, yet the model answers
    \textbf{``No''} to the question \textit{``Is there a frisbee in the image?''}. Perturbation-only
    contrastive methods further suppress the already weak frisbee region, and—because
    they operate in a single destructive path—VCD-style methods cannot recover from
    such loss of evidence, again causing the model to deny its presence.
\end{itemize}

Overcoming this phenomenon requires a mechanism that can dynamically and reliably
intervene during decoding. Such an approach must continually steer the model
toward visually grounded predictions and prevent it from reverting to descriptions
dominated by an incorrect language prior \cite{yang2025mitigating, zhu2025ibd}.
As illustrated in \cref{fig:intro}(c), our method effectively resolves this imbalance.
To truly break the illusion of understanding, we introduce \textbf{Positive-and-Negative Decoding (PND)}, a training-free, plug-and-play framework that performs real-time \textbf{Bayesian belief adjustment} during inference. PND achieves this by injecting a carefully designed dual-path contrast mechanism \cite{chopin2015some} that dynamically re-balances the prior and likelihood.

\begin{enumerate}[label=\textbf{\arabic*.}, leftmargin=*, itemsep=2pt, topsep=2pt]

    \item \textbf{The Positive Pathway (Amplifying the Likelihood):}
    This pathway uses multi-layer cross-modal attention to gather relevant visual evidence and amplify it. 
    By strengthening high-salience visual features, it boosts the \textbf{visual likelihood} and steers the model toward grounded, faithful descriptions.

    \item \textbf{The Negative Pathway (Isolating the Prior):}
    This pathway builds a targeted counterfactual by identifying the core evidence regions and selectively degrading them. By removing only the minimal visual cues the model still relies on, it induces an ``evidence-blind'' state while preserving useful prior information, leading the model to rely more heavily on its \textbf{language prior}.  
The resulting output reveals the model’s underlying, prior-driven hallucination tendencies.

\end{enumerate}

The "meeting" of these pathways is orchestrated by our decoding objective. By contrasting the model's outputs from the enhanced-likelihood (Positive) path against the prior-dominant (Negative) path, PND applies symmetrical pressure. It steers the generation trajectory towards object-level truth (high visual likelihood) and away from misleading contextual beliefs (dominant language prior), thus resolving the Bayesian imbalance that leads to hallucinations \cite{dathathriplug, zhu2025ibd}. Our contributions can be summarized as follows:
\begin{itemize}[itemsep=2pt,topsep=2pt]
    \item We present PND, a training-free, plug-and-play decoding framework that uses a \textbf{dual-path contrastive mechanism} to suppress hallucinations during inference.
    \item Our method uniquely leverages multi-layer attention to dynamically generate positive (likelihood-amplifying) and negative (prior-isolating) guidance, implementing a robust Bayesian belief adjustment at inference time.
    \item Extensive experiments across multiple VLMs and benchmark datasets demonstrate that PND achieves state-of-the-art performance in suppressing object hallucinations, significantly outperforming existing methods.
\end{itemize}

\section{Related Work}
\subsection{Vision-Language Models (VLMs)}
The current paradigm of Large Multimodal Models (LMMs) enhances powerful LLMs with visual understanding. This is typically achieved by connecting a pre-trained vision encoder \cite{dosovitskiy2020image} to the LLM via a lightweight adapter \cite{liu2023visual, liu2024improved}. This architecture, particularly when refined with large-scale \textbf{visual instruction tuning} \cite{liu2023visual}, has proven highly effective. It has produced a wave of influential \textbf{open-source models}—such as LLaVA-1.5 \cite{liu2024improved}, Qwen-VL \cite{bai2023qwenvlversatilevisionlanguagemodel}, and Deepseek-VL \cite{lu2024deepseek}—as well as cutting-edge proprietary systems like OpenAI's GPT models and Google's Gemini \cite{team2023gemini}. These models demonstrate unprecedented conversational skills and performance \cite{hu2024bliva}.

While this design enables powerful fluency, it also causes the models to inherit the vast parametric knowledge of their underlying LLM. This reliance substantially contributes to the \textbf{Bayesian imbalance} we address: their strong \textbf{language priors} can easily override factual visual evidence, leading to the pervasive problem of hallucination \cite{liu2025reducing, gunjal2024detecting, damonlpsg2023vcd, zhu2025ibd}.

\subsection{Hallucination in Vision-Language Models}

Hallucination in VLMs denotes generated content inconsistent with visual input \cite{zhong2024investigating}.
Object hallucination—false positives (describing non-existent objects) or false negatives (omitting present ones)—is the most studied and practically significant form \cite{gunjal2024detecting, bai2024hallucination}. These errors are closely linked to strong language priors \cite{peng2024event} and attention misalignment, where models rely on contextual cues instead of object-level evidence \cite{shu2025large}. Existing mitigation strategies fall into two categories.
Training-based methods modify parameters through RLHF \cite{ouyang2022training, huang2024opera}, curated datasets \cite{bai2024hallucination}, or architectural changes \cite{an2025mitigating}. Although effective, they are computationally costly and often degrade other multimodal abilities. Inference-time methods provide a practical alternative.
Recent approaches such as Visual Contrastive Decoding (VCD) \cite{damonlpsg2023vcd}, AGLA \cite{an2025mitigating}), and VAF \cite{yin2025clearsight} contrast predictions under perturbed visual inputs to detect prior-dominant tokens. From a Bayesian perspective, these techniques perform single-path perturbation control: they weaken visual likelihood and down-weight tokens that remain unchanged. Despite their effectiveness, this mechanism is inherently one-sided; it neither amplifies evidence-bearing regions nor separates the influence of the language prior.

\begin{figure*}[t]
\centering
\includegraphics[width=0.85\textwidth]{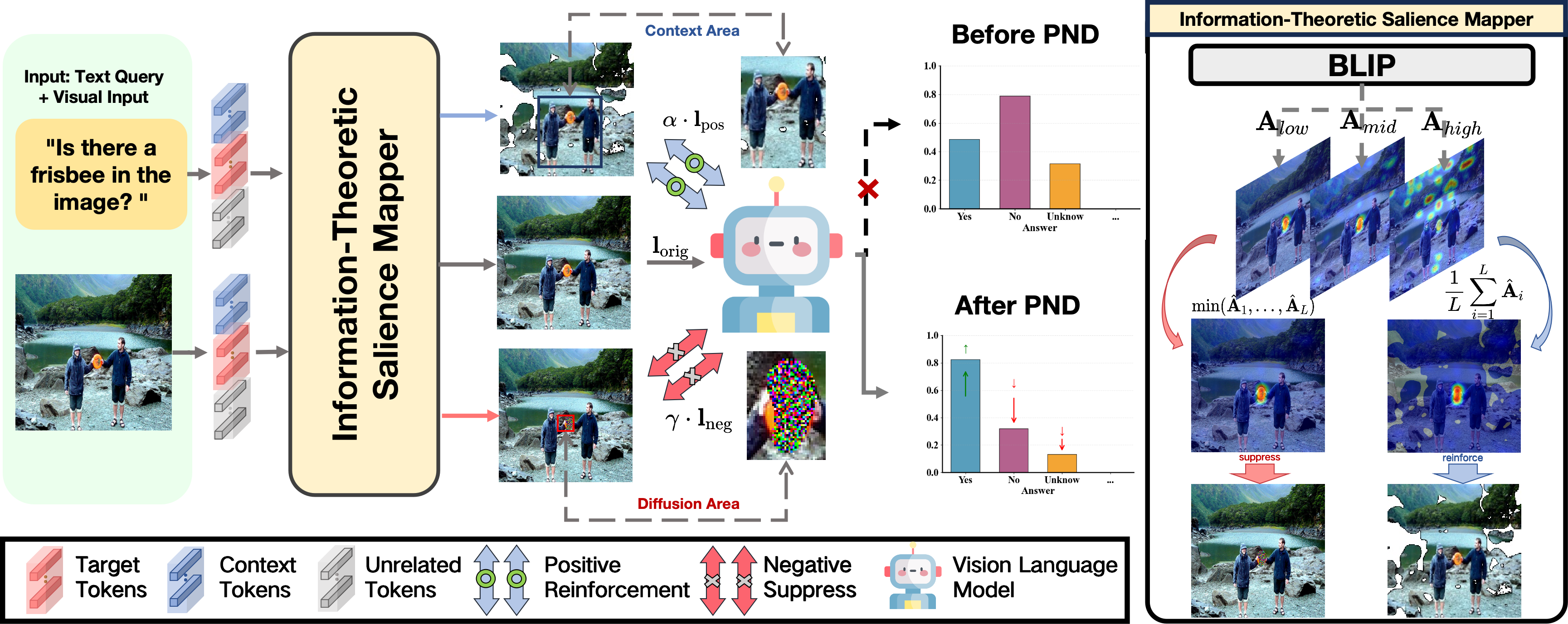} 
\caption{\textbf{Overview of the PND framework for belief-adjusted decoding.}
Given an input image, we first extract multi-layer cross-modal attention maps to estimate query-aligned visual evidence. These maps guide the construction of two perturbed visual representations: a \emph{positive} view $\mathbf{V}_{\mathrm{pos}}$ that amplifies evidence, and a \emph{negative} view $\mathbf{V}_{\mathrm{neg}}$ that suppresses it.  Passing each view through the VLM yields three logits (original, positive, and negative), whose contrast reveals whether a token is driven by visual likelihood or by the language prior. The final next-token probability is obtained by a belief-adjusted combination of these logits, enabling the model to recover visually grounded predictions and reduce hallucination.}
\label{P-2}
\end{figure*}

Our approach offers a complementary, structured perspective.
Rather than relying on a single perturbed view, PND introduces a dual-path formulation probing both belief sources. The negative path builds a controlled counterfactual removing multi-layer consensus evidence to approximate the prior, while the positive path reinforces salient regions via attention-guided enhancement.
This design treats CAM as a differentiable, architecture-agnostic proxy separating likelihood- and prior-dominant regions, enabling principled, model-agnostic belief-adjusted decoding. The symmetric formulation provides a clearer approximation of Bayesian decomposition than perturbation-only methods.


\section{Method}
\label{sec:method}
This section introduces PND, our inference-time framework guided by Bayesian belief adjustment. Rather than assuming a fully parametric Bayesian model, we use this viewpoint to describe an empirically observed imbalance: modern VLMs rely heavily on linguistic self-consistency while progressively under-utilizing visual evidence in deeper decoding layers (see \cref{fig:attention-allocation}). Under this imbalance, PND aims to dynamically re-weight the language prior and visual likelihood during token generation.

To achieve this, PND contrasts the model’s behavior under two visual representations. A positive representation $\mathbf{V}_{\mathrm{pos}}$ amplifies salient visual evidence, while a negative representation $\mathbf{V}_{\mathrm{neg}}$ attenuates or removes such evidence to isolate the model’s language prior. The premise is simple: tokens dominated by the prior remain insensitive to visual perturbations, whereas likelihood-driven tokens exhibit strong shifts. As illustrated in \cref{P-2}, our framework consists of two components: attention-derived salience maps for constructing $\mathbf{V}_{\mathrm{pos}}$ and $\mathbf{V}_{\mathrm{neg}}$, and a belief-adjusted decoding objective that integrates logits from all three paths into a single next-token distribution.

\subsection{Disentangling Evidence and Context via Attention}
\label{sec:method_attention}

We revisit multimodal decoding through a conceptual Bayesian lens, where the next-token distribution is jointly influenced by linguistic expectations and dynamically evolving image-derived evidence:

\begin{equation}
    p(y \mid x_v, x_t) \,\propto\,
    \underbrace{p(y \mid x_t)}_{\text{language prior}}
    \cdot
    \underbrace{p(x_v \mid y)}_{\text{visual likelihood}} .
\label{eq:bayes}
\end{equation}
While this factorization usefully interprets hallucination, explicitly decomposing a VLM’s hidden features into these components is intractable. Instead, we seek a practical proxy aligning with observable model behavior \cite{listreamlining, cai2008comment}.

\paragraph{Cross-modal attention as an empirical proxy.}
We interpret visual embedding $\mathbf{V}$ as comprising evidence-bearing $\mathbf{V}_{\mathrm{evidence}}$ (object features supporting likelihood) and contextual $\mathbf{V}_{\mathrm{context}}$ (semantics reinforcing language priors). Hallucinations emerge when models overweight $\mathbf{V}_{\mathrm{context}}$ and underweight $\mathbf{V}_{\mathrm{evidence}}$ during decoding. Despite the early-aggregation hypothesis (visual evidence integrating early for indirect later access), deeper layers increasingly favor language priors over direct visual evidence. Thus, observed attention decline empirically indicates reduced direct visual grounding rather than definitive information loss.

To approximate this decomposition, we extract cross-modal attention maps from an external vision–language model (BLIP-ITM \cite{li2022blip}). These maps quantify relevance between textual queries $\mathbf{Q}_{\text{text}}$ and visual patches $\mathbf{K}_{\text{vis}}$, estimating where visual evidence resides. Given textual queries and visual keys in layer $i$, attention map $\mathbf{A}_i$ is

\begin{equation}
\mathbf{A}_i=\mathrm{softmax}\!\left(
\mathbf{Q}_{\text{text}}\!\left(\mathbf{K}_{\text{vis}}^{(i)}\right)^{\!\top} / \sqrt{d_k}
\right).
\label{eq:attention}
\end{equation}

\begin{figure}[t]
\centering
\includegraphics[width=0.9\columnwidth]{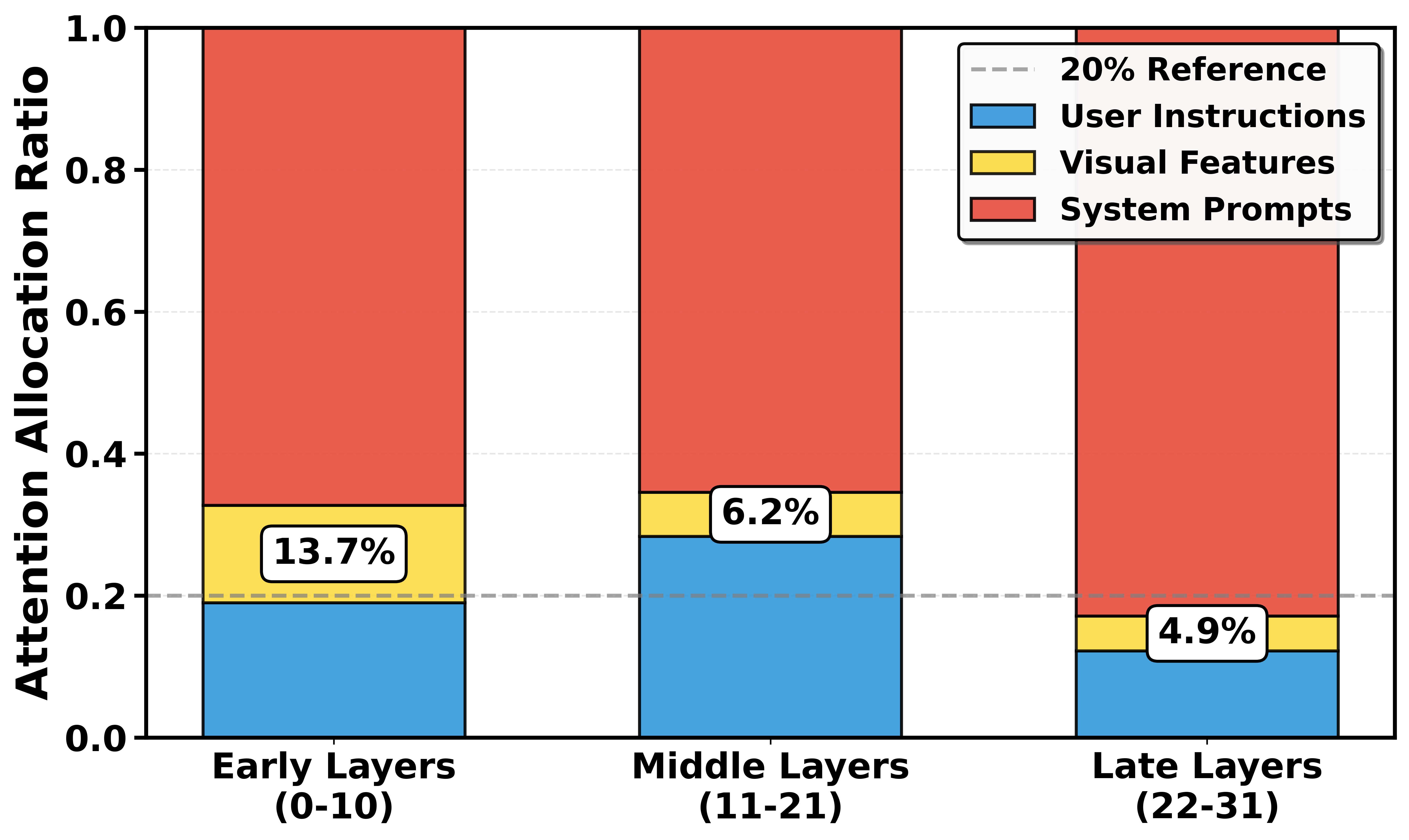} 
\caption{\textbf{Empirical evidence for Bayesian imbalance in multimodal decoding.} We plot the layer-wise allocation of cross-modal attention in a representative VLM. Early layers attend to visual evidence, but deeper layers shift attentional most entirely toward textual context, reflecting the accumulation of a strong language prior. This progressive decline in visual contribution indicates that $p(x_v \mid y)$ is underweighted relative to $p(y \mid x_t)$ during token generation, providing a direct motivation for our \emph{Bayesian belief adjustment} design in PND.}
\label{fig:attention-allocation}
\end{figure}

\paragraph{Layerwise distinction between likelihood and prior.}
CAMs across layers exhibit systematic behavior: early layers emphasize fine-grained object regions, whereas deeper layers shift toward global semantics. Empirically (\cref{fig:attention-allocation}), visual patches receive substantially less attention in deeper layers, dominated by user instructions and system prompts. This trend aligns with our Bayesian interpretation: visual likelihood fades with depth, while language priors accumulate. Motivated by this, we fuse multi-layer CAMs $\{\mathbf{A}_1, \dots, \mathbf{A}_L\}$ into salience maps that highlight evidence-bearing regions while suppressing context-driven ones. These fused maps provide a differentiable, architecture-agnostic proxy for separating likelihood-dominant and prior-dominant regions, forming the foundation of the positive and negative visual pathways. Supplementary~Material~IV details individual layer contributions.

\subsection{Positive and Negative Visual Augmentation}

With cross-modal attention maps providing a practical proxy for disentangling
\emph{evidence-bearing} and \emph{contextual} regions (\cref{sec:method_attention}),
we now describe how these signals are operationalized to construct the two
visual pathways used for belief adjustment. As illustrated in \cref{P-2}, PND generates two complementary visual representations:
a \textbf{positive} view $\mathbf{V}_{\mathrm{pos}}$ that \emph{explicitly amplifies the
visual likelihood}, and a \textbf{negative} view $\mathbf{V}_{\mathrm{neg}}$
that \emph{strategically suppresses evidence to expose the language prior}.

The goal of these augmentations is not to alter the semantics of the image,
but to selectively modulate the strength of the model’s visual cues. When the
model processes $\mathbf{V}_{\mathrm{pos}}$, regions identified as
high-evidence receive proportionally greater emphasis, increasing the model’s
sensitivity to grounded signals. Conversely, $\mathbf{V}_{\mathrm{neg}}$
reduces or removes these same regions—via attention-guided degradation—to
approximate a counterfactual setting in which visual evidence is attenuated.
Tokens that genuinely rely on $p(x_v \mid y)$ exhibit strong shifts across the
two views, whereas hallucinated (prior-dominant) tokens remain comparatively
stable.

These augmented representations form the foundation for our dual-path
decoding strategy, enabling PND to contrast likelihood-sensitive and
prior-sensitive behavior at inference time. Details of the specific
construction procedures for $\mathbf{V}_{\mathrm{pos}}$ and
$\mathbf{V}_{\mathrm{neg}}$ follow in the next sections.

\subsubsection{Positive Enhancement: Amplifying the visual likelihood}

The positive pathway is designed to counteract the empirical 
\textbf{Bayesian imbalance} revealed in our attention analysis 
(\cref{fig:attention-allocation}). 
Across layers, cross-modal attention assigns only a small portion of its 
budget to visual patches (13.7\% in early layers, decreasing to 6.2\% and 
4.9\% in middle and late layers), while textual context---such as \textbf{User 
Instructions} and \textbf{System Prompts}---dominates. 
This attenuation of visually grounded cues indicates that the 
\textbf{visual likelihood} $p(x_v \mid y)$ is systematically underweighted, 
motivating an intervention that selectively strengthens evidence-bearing 
regions. We compute a token-level evidence weight from multi-layer cross-modal attention, following the detailed formulation provided in the Supplementary~Material~VII. To implement this correction, we construct a salience map by aggregating 
multi-layer cross-modal attention, following attention-fusion strategies 
explored in prior work \cite{tang2024attention, li2023multi}:
\begin{equation}
    \mathbf{M}_{\mathrm{fused}}
    = \frac{1}{L} \sum_{i=1}^{L} \hat{\mathbf{A}}_i ,
\end{equation}
where $\hat{\mathbf{A}}_i$ is the normalized attention map at layer $i$.  
This fused map highlights the regions the model implicitly associates with the 
query and thus approximates the \textit{evidence} component in our Bayesian 
interpretation. We then amplify these regions using a multiplicative modulation inspired by 
feature-boosting mechanisms in prior work \cite{poudel2021deep}:
\begin{equation}
    \mathbf{V}_{\mathrm{pos}}
    = \mathbf{V}_{\mathrm{orig}}
      \odot \left( 1 + \lambda \cdot \mathbf{M}_{\mathrm{fused}} \right),
\end{equation}
where $\lambda$ controls the strength of amplification and $\odot$ denotes 
element-wise scaling.  
This operation does not alter the semantics of the image; instead, it gradually increases 
the relative prominence of evidence-bearing features, thereby encouraging the model to 
more faithfully reflect the \textbf{visual likelihood} during decoding.

\subsubsection{Negative Degradation: Isolating the language prior}

The negative pathway constructs a counterfactual visual input that \emph{carefully isolates the language prior} without destroying the useful parts of it. As shown in \cref{fig:attention-allocation}, modern VLMs already allocate very little deep-layer attention to visual patches (e.g., 4.9\%), making global perturbations wasteful and prone to erasing helpful priors. Instead, we remove only the \emph{minimal evidence} identified by multi-layer CAM consensus—preserving most visual information while still inducing strongly maximal hallucination. This subtle intervention forces decoding to rely almost entirely on $p(y \mid x_t)$ and reveals the model’s prior-driven bias.

\paragraph{Identifying minimal evidence via attention consensus.}
Following insights from evidence localization \cite{shi2023towards}, we compute
a \emph{consensus map} that highlights visual regions consistently attended to
across layers.  
Specifically, we take the pixel-wise minimum over normalized attention maps:
\begin{equation}
    \mathbf{M}_{\mathrm{consensus}}
    = \min(\hat{\mathbf{A}}_1, \ldots, \hat{\mathbf{A}}_L).
\end{equation}
This soft intersection yields a conservative estimate of the
evidence-bearing regions that contribute to the visual likelihood.
A binary mask is then produced via thresholding:
\begin{equation}
    \mathbf{M}_{\mathrm{mask}} = \mathbb{I}[\mathbf{M}_{\mathrm{consensus}} \ge \tau].
\end{equation}

\paragraph{Semantic degradation through DDPM forward process.}
To meaningfully degrade these regions without introducing artifacts, we employ the forward noising process of a DDPM \cite{ho2020denoising}, which produces corrupted features that remain structurally coherent. Unlike Gaussian noise, DDPM corruption yields distributionally plausible and semantically aligned patterns, avoiding the out-of-distribution artifacts that models tend to ignore. For noise level $T$, the corrupted representation is sampled as:

\begin{equation}
    \mathbf{V}_{\mathrm{noise}}
    =
    \sqrt{\bar{\alpha}_T}\, \mathbf{V}_{\mathrm{orig}}
    +
    \sqrt{1 - \bar{\alpha}_T}\, \boldsymbol{\epsilon},
    \quad
    \boldsymbol{\epsilon} \sim \mathcal{N}(0, \mathbf{I}),
\end{equation}
where $\bar{\alpha}_T$ controls the resulting signal-to-noise ratio.

\paragraph{Constructing the negative counterfactual.}
The final negative input is assembled by carefully replacing only the masked
evidence-bearing regions with the corresponding DDPM-corrupted features.

\begin{equation}
    \mathbf{V}_{\mathrm{neg}}
    =
    \mathbf{V}_{\mathrm{orig}} \odot (1 - \mathbf{M}_{\mathrm{mask}})
    +
    \mathbf{V}_{\mathrm{noise}} \odot \mathbf{M}_{\mathrm{mask}}.
\end{equation}
This targeted degradation removes the model’s remaining access to visual
likelihood while preserving overall image statistics, yielding a
prior-isolating counterfactual.  
Such counterfactual inputs have been shown to expose hallucination tendencies in
VLMs \cite{chen2024multi, guan2024hallusionbench}, and here they form the
negative component of our dual-path decoding framework.

\subsection{PND: Decoding as Bayesian Belief Adjustment}

With $\mathbf{V}_{\mathrm{orig}}$, $\mathbf{V}_{\mathrm{pos}}$, and 
$\mathbf{V}_{\mathrm{neg}}$ prepared, we now describe the decoding step of PND in detail.
This stage performs a lightweight, inference-only \textbf{Bayesian belief
adjustment}, dynamically re-balancing the relative influence of the \textbf{language prior} and 
\textbf{visual likelihood} at every token.

We run three parallel forward passes to obtain logits  
$\mathbf{l}_{\mathrm{orig}}$,  
$\mathbf{l}_{\mathrm{pos}}$ (likelihood-amplified),  
and $\mathbf{l}_{\mathrm{neg}}$ (prior-isolated).  
PND combines these signals through a contrastive update:
\begin{equation}
    \mathbf{l}_{\mathrm{PND}}
    =
    \mathbf{l}_{\mathrm{orig}}
    + \alpha \, \mathbf{l}_{\mathrm{pos}}
    - \gamma \, \mathbf{l}_{\mathrm{neg}},
    \label{eq:pnd_control_law}
\end{equation}
where $\alpha, \gamma \ge 0$ are balancing coefficients.
The positive term boosts visually grounded candidates, while the negative term
suppresses prior-driven tokens that persist without visual support, consistent
with findings in prior hallucination-control work
\cite{tang2024attention, guan2024hallusionbench}. To avoid introducing implausible candidates, we retain only tokens that are
credible under the model’s original distribution:
\begin{equation}
    \mathbf{l}_{\mathrm{final}}
    =
    \mathbf{l}_{\mathrm{PND}}
    \odot
    \mathbb{I}\!\left[
        \mathbf{l}_{\mathrm{orig}}
        \ge
        \log(\beta) + \max(\mathbf{l}_{\mathrm{orig}})
    \right],
\end{equation}
where $\beta$ is a confidence threshold.  
The final probability distribution is obtained by applying softmax to
$\mathbf{l}_{\mathrm{final}}$.

\begin{table*}[t]
\centering
\caption{Performance comparison on the POPE benchmark, evaluated on LLaVA1.5-7B, InstructBLIP-7B, and other models. Our PND framework shows \textbf{state-of-the-art performance}, with superior Accuracy and F1-scores across most settings compared to the baseline and competing methods (VCD \cite{damonlpsg2023vcd}, VAF \cite{yin2025clearsight}, AGLA \cite{an2025mitigating}).}
\setlength{\tabcolsep}{12pt} 
\renewcommand{\arraystretch}{0.7} 
\label{tab1}
\begin{tabular}{@{}llccccc@{}}
\toprule
\textbf{Model} & \textbf{Category} & \textbf{Methods} & \textbf{Accuracy} & \textbf{Precision} & \textbf{Recall} & \textbf{F1} \\
\midrule
& & regular & 78.53 & 82.77 & 72.06 & 77.04 \\
& & VCD & 81.13 & 84.52 & 76.72 & 80.43 \\
& & VAF & 80.88 & 86.84 & 76.58 & 81.39 \\
& & AGLA & 83.13 & 86.97 & \textbf{77.95} & 82.21 \\
& \multirow{-5}{*}{adversarial} & \textbf{PND(ours)} & \textbf{84.03}  & \textbf{90.55} & 77.43 & \textbf{83.48}\\
\cmidrule(lr){2-7}
& & regular & 81.56 & 94.14 & 77.25 & 84.87 \\
& & VCD & 84.89 & 93.86 & 77.92 & 85.16 \\
& & VAF & 84.50 & 95.76 & 78.85 & 86.49 \\
& & AGLA & 85.21 & 94.92 & \textbf{80.87} & 87.34 \\
& \multirow{-5}{*}{popular} & \textbf{PND(ours)} & \textbf{86.10}  & \textbf{98.41} & 80.87 & \textbf{88.79} \\
\cmidrule(lr){2-7}
& & regular & 83.00 & 96.23 & 82.99 & 89.12 \\
& & VCD & 86.81 & 96.11 & 85.38 & 90.43 \\
& & VAF & 85.68 & 95.33 & 88.65 & 91.87 \\
& & AGLA & 87.12 & 96.42 & \textbf{88.78} & 92.44 \\
\multirow{-15}{*}{\textbf{LLaVA1.5-7B \cite{liu2024improved}}} & \multirow{-5}{*}{random} & \textbf{PND(ours)} & \textbf{87.33} & \textbf{98.29} & 88.99 & \textbf{93.41}  \\
\midrule
& & regular & 75.66 & 74.09 & 78.93 & 76.43 \\
& & VCD & 78.43 & 79.04 & 79.92 & 79.48 \\
& & VAF & 78.84 & 78.49 & 79.47 & 78.98 \\
& & AGLA & 81.13 & 82.35 & 80.74 & 81.54 \\
& \multirow{-5}{*}{adversarial} & \textbf{PND(ours)} & \textbf{82.20}  & \textbf{82.99} & \textbf{81.00} & \textbf{81.98} \\
\cmidrule(lr){2-7}
& & regular & 78.00 & 77.30 & 79.26 & 78.27 \\
& & VCD & 82.24 & 82.56 & 81.02 & 81.79 \\
& & VAF & 81.72 & 82.54 & 80.20 & 81.36 \\
& & AGLA & 83.21 & 85.96 & \textbf{80.47} & 83.12 \\
& \multirow{-5}{*}{popular} & \textbf{PND(ours)} & \textbf{84.83}  & \textbf{87.83} & 80.86 & \textbf{84.20} \\
\cmidrule(lr){2-7}
& & regular & 81.10 & 82.41 & 79.06 & 80.70 \\
& & VCD & 85.12 & 88.57 & 80.20 & 84.18 \\
& & VAF & 84.89 & 88.74 & 79.71 & 83.98 \\
& & AGLA & 86.31 & 92.32 & 79.72 & 85.56 \\
\multirow{-15}{*}{\textbf{InstructBLIP-7B \cite{dai2023instructblip}}} & \multirow{-5}{*}{random} & \textbf{PND(ours)} & \textbf{87.63}  & \textbf{93.52} & \textbf{80.86} & \textbf{86.73}  \\
\midrule
\multicolumn{7}{c}{\textit{Other Models}} \\
\midrule
\multirow{2}{*}{\textbf{LLAVA1.5-13B \cite{liu2024improved}}} & \multirow{2}{*}{adversarial} & regular   & 79.80 & 83.60 & 74.13 & 78.58 \\
&                              & \textbf{PND(ours)} &    \textbf{84.96}  & \textbf{91.99} & \textbf{76.60}  & \textbf{83.59}   \\
\multirow{2}{*}{\textbf{InstructBLIP-13B \cite{dai2023instructblip}}} & \multirow{2}{*}{adversarial} & regular   &  76.03 & 76.22 & 75.66 & 75.94 \\
&                              & \textbf{PND(ours)} &    \textbf{82.83}  & \textbf{84.75} & \textbf{80.06} & \textbf{82.34}  \\
\multirow{2}{*}{\textbf{QwenVL-7B \cite{bai2023qwenvlversatilevisionlanguagemodel}}}    & \multirow{2}{*}{adversarial} & regular   & 80.96 & 91.06 & 68.66 & 78.29 \\
&                              & \textbf{PND(ours)} &  \textbf{82.46}      &  \textbf{93.55}     &   \textbf{69.73}    &   \textbf{79.90}    \\
\multirow{2}{*}{\textbf{Qwen3VL-2B \cite{yang2025qwen3technicalreport}}} & \multirow{2}{*}{adversarial} & regular   &  86.53     &   86.29    &   86.86    &   86.57    \\
&                              & \textbf{PND(ours)} &     \textbf{87.26}  &   \textbf{88.02}    &   86.26    &  \textbf{87.13}     \\
\multirow{2}{*}{\textbf{InternVL2-2B \cite{wang2024enhancing}}} & \multirow{2}{*}{adversarial} & regular   & 81.06 & 85.35 & 75.00 & 79.84 \\
&                              & \textbf{PND(ours)} &    \textbf{83.33}   & \textbf{88.22}      &   \textbf{76.93}    &   \textbf{82.19}   \\
\bottomrule
\end{tabular}
\end{table*}
\begin{table*}[t]
\centering
\caption{Performance comparison of different hallucination mitigation methods on the MME benchmark}
\label{tab2}
\setlength{\tabcolsep}{12pt} 
\renewcommand{\arraystretch}{0.70} 
\begin{tabular}{@{}clccccc@{}}
\toprule
\multirow{2}{*}{\textbf{Model}} &
  \multirow{2}{*}{\textbf{Method}} &
  \multicolumn{2}{c}{\textbf{Object-level}} &
  \multicolumn{2}{c}{\textbf{Attribute-level}} &
  \multirow{2}{*}{\textbf{Total Score}} \\
\cmidrule(lr){3-4} \cmidrule(lr){5-6}
                                   &  & Existence       & Count           & Position        & Color           &  \\
\midrule
\multirow{5}{*}{\textbf{LLaVA1.5-7B}} & regular              & 160.00          & 121.67          & 105.00          & 145.00          & 531.67               \\
                                   & VCD                  & 185.00          & 125.33          & 105.00          & 150.00          & 565.33               \\
                                   & AGLA                 & \textbf{195.00}          & 138.33          & 120.33          & 155.00          & 608.67               \\
                                   & VAF                  & \textbf{195.00}          & 133.67          & 105.00          & 150.00          & 583.67               \\
                                   & \textbf{PND(ours)}         & \textbf{195.00}  & \textbf{143.33}  & \textbf{123.33}  & \textbf{160.00}  & \textbf{621.67}      \\
\midrule
\multirow{5}{*}{\textbf{InstructBLIP-7B}}    & regular              & 170.00          & 50.00           & 53.33           & 113.33          & 386.66               \\
                                   & VCD                  & 170.00          & 50.00           & 53.33           & 114.33          & 387.67               \\
                                   & AGLA                 & 175.00          & 55.00           & 58.33           & 118.33          & 406.67               \\
                                   & VAF                  & 175.00          & 51.67           & 57.33           & 113.33          & 397.33               \\
                                   & \textbf{PND(ours)}         & \textbf{180.00}  & \textbf{60.00}   & \textbf{63.33}   & \textbf{120.00}  & \textbf{423.33}   \\

\midrule
\multicolumn{7}{c}{\textit{Other Models}} \\
\midrule
\multirow{2}{*}{\textbf{QwenVL-7B}}    & regular              & 150.00          & 145.00           & 113.33          & 160.00          &       568.33         \\
                                   & \textbf{PND(ours)}         & \textbf{170.00}  & \textbf{150.00}   & \textbf{123.33}   & \textbf{173.33}  & \textbf{616.67}   \\
\multirow{2}{*}{\textbf{Qwen3VL-2B}}    & regular              &      190.00     &   155.00        &    143.33       &     150.00      &     638.33          \\
                                   & \textbf{PND(ours)}         & 190.00  & \textbf{160.00}   & \textbf{158.33}   & \textbf{160.00}  & \textbf{668.33}   \\
\multirow{2}{*}{\textbf{InternVL2-2B}}    & regular              &   168.33        &   88.33        &     103.33      &      90.00     &         450.00      \\
                                   & \textbf{PND(ours)}         & \textbf{185.00}  & \textbf{93.33}   & \textbf{141.67}   & \textbf{110.00}  & \textbf{530.00}   \\

\bottomrule
\end{tabular}
\end{table*}

\begin{table}[t]
\centering
\caption{Performance comparison of different hallucination mitigation methods on the CHAIR benchmark}
\setlength{\tabcolsep}{2pt} 
\renewcommand{\arraystretch}{0.7} 
\label{tab3}
\begin{tabular}{@{}cllll@{}}
\toprule
\textbf{Model} & \textbf{Methods} & \textbf{$\mathcal{C}$s} $\downarrow$ & \textbf{$\mathcal{C}$i} $\downarrow$ & \textbf{Recall} $\uparrow$ \\
\midrule
\multirow{5}{*}{\textbf{LLaVA1.5-7B}} & regular    & 51.0   & 17.6   & 74.4   \\
                                & VCD        & 51.0   & 16.7   & 77.2   \\
                                & AGLA       & 47.0   & 14.2   & 77.8   \\
                                & VAF & 53.0   & 16.5   & 76.9   \\
                                & \textbf{PND(ours)} & \textbf{46.0}  & \textbf{14.0}  & \textbf{78.1}  \\
\midrule
\multirow{5}{*}{\textbf{InstructBLIP-7B}} & regular    & 58.0   & 16.3   & 71.1   \\
                                       & VCD        & 59.0   & 14.8   & 72.0   \\
                                       & AGLA       & 46.0   & 12.3   & 71.5   \\
                                       & VAF & 56.0   & 15.1   & 70.4   \\
                                       & \textbf{PND(ours)} & \textbf{42.0}  & \textbf{11.2}  & \textbf{72.1} \\
\bottomrule
\end{tabular}
\end{table}

\section{Experiments}
\subsection{Experiments Settings}

\paragraph{Datasets}
To avoid benchmark-driven evaluation, we adopt a \emph{structured,
multi-level} protocol that examines hallucination from complementary
perspectives rather than relying on a single metric.

\begin{itemize}[leftmargin=*]

    \item \textbf{POPE \cite{li2023evaluating}.}  
    Probes \emph{object-level} hallucination via a Yes/No formulation.
    We evaluate on MSCOCO \cite{lin2014microsoft}, A-OKVQA
    \cite{schwenk2022okvqa}, and GQA \cite{hudson2019gqa}, which cover random,
    popular, and adversarial sampling strategies.

    \item \textbf{MME \cite{yin2024survey}.}  
    Measures \emph{perceptual and attribute-level} capabilities across more than
    ten dimensions.  
    We use MME primarily to assess whether PND reduces hallucination \emph{without
    harming} standard multimodal competence.

    \item \textbf{CHAIR \cite{rohrbach2018object}.}  
    Evaluates hallucination in \emph{open-ended captioning} by comparing object
    nouns in generated captions to ground-truth sets.  
    We report CHAIR$_s$ and CHAIR$_i$.

    \item \textbf{GCCCE (ours).}  
    Targets \emph{high-level semantic and commonsense} consistency.  
    Using GPT-4.1 as a judge, GCCCE scores responses on Relevancy, Accuracy,
    Common Sense, and Fine-grained Precision.

\end{itemize}

Together, these four benchmarks form a concise evaluation suite covering object grounding, perceptual attributes, open-ended generation, and semantic coherence. They enable multi-perspective hallucination assessment across perception and reasoning.
\paragraph{MLLM Backbones}

To demonstrate PND's generality, we evaluate it on four widely used
open-source MLLMs—\textbf{LLaVA}, \textbf{InstructBLIP}, \textbf{InternVL},
and \textbf{Qwen-VL}. These models span vision–language fusion designs from projection-based alignment to query-driven transformer architectures.  
Consistent gains across heterogeneous systems suggest PND acts as a model-agnostic decoding enhancement rather than an architecture-specific trick. Our method trades modest inference overhead for improved visual grounding, mainly from BLIP-based attention extraction. PND adds small decoding-time overhead, far lighter than training-based mitigation, and can be enabled when higher reliability is needed. Detailed efficiency analysis appears in Supplementary~Material~VI.

\subsection{Experiments Results}

\paragraph{Results on POPE}
Our primary evaluation on the POPE benchmark (\cref{tab1}) confirms PND's efficacy in suppressing object hallucination. Our framework yields substantial gains across LLaVA-1.5, InstructBLIP, and Qwen-VL, achieving an average improvement of \textbf{6.4\%} in Accuracy and \textbf{5.5\%} in F1-score over the greedy decoding baseline. Crucially, the performance gains are most pronounced on the challenging popular and adversarial subsets. This directly validates our Bayesian hypothesis: these subsets are specifically designed to trigger hallucinations via powerful, yet fallacious, linguistic priors. PND's success here indicates it effectively counteracts this over-weighted \textbf{language prior}, forcing the model to rely on the \textbf{visual likelihood}. Detailed robustness and scalability analyses, including 7B/13B variants, are in Supplementary~Material~II.

\paragraph{Results on MME}
To assess PND's impact on broader multimodal capabilities, we evaluated its performance on the comprehensive MME benchmark (\cref{tab2}). Our framework establishes a new state-of-the-art (SOTA) across the evaluated perception sub-tasks. Substantial improvements are observed not only in object-level perception (\textit{Existence}, \textit{Count}) but also, crucially, in fine-grained attribute grounding (\textit{Position}, \textit{Color}). This demonstrates that PND functions as a holistic enhancement for visual fidelity, improving a wide spectrum of perception skills rather than being a narrow fix with potential side effects. Complete MME results are in Supplementary~Material~III.
\begin{figure}[t]
\centering
\includegraphics[width=0.42\textwidth]{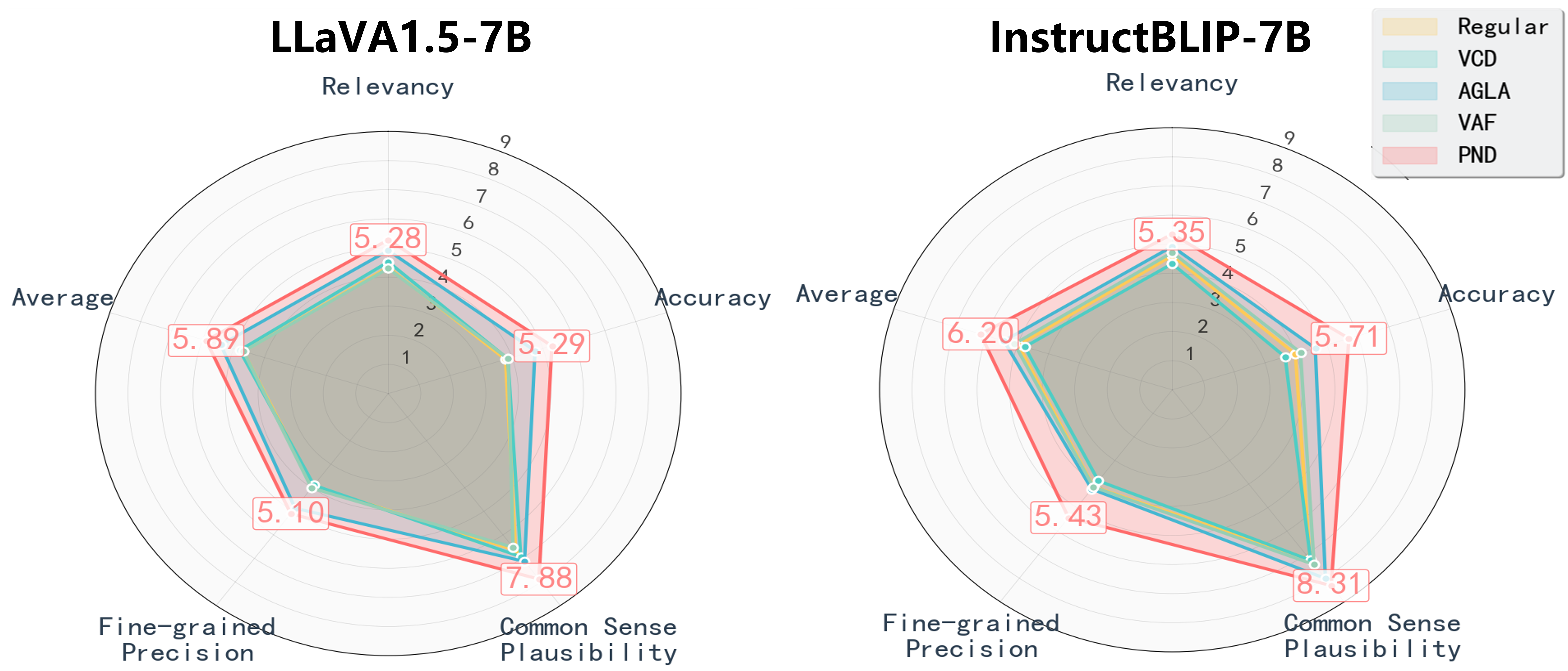} 
\caption{Performance comparison of different hallucination
mitigation methods on the GCCCE benchmark.}
\label{P-GCCCC}
\end{figure}

\begin{figure*}[t]
\centering
\includegraphics[width=0.80\textwidth]{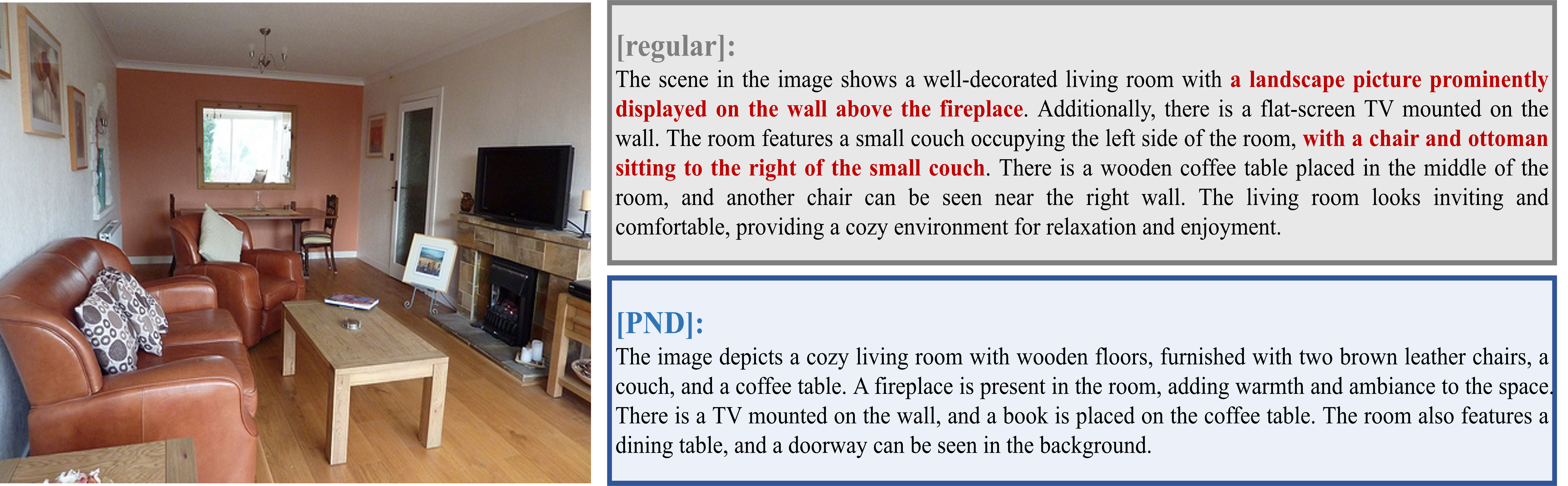} 
\caption{Qualitative comparison for "describe the scene captured in the image". The baseline model ([regular]) produces a description with significant factual errors (inventing an 'ottoman'), which are highlighted in \textcolor{red}{red}. In contrast, our PND-enhanced model successfully suppresses these errors and generates a visually faithful and accurate description.}
\label{P-case}
\end{figure*}

\paragraph{Results on CHAIR}
To validate PND beyond discriminative, polling-based tasks, we evaluated its performance on open-ended caption generation using the CHAIR benchmark (\cref{tab3}). Our method significantly reduces object hallucination, evidenced by the sharp drop in both CHAIR$_s$ and CHAIR$_i$ scores for tested backbones. This result directly validates the efficacy of our negative pathway in penalizing object tokens that lack sufficient visual grounding.

\paragraph{Results on GCCCE}

To assess higher-level cognitive effects, we apply GCCCE to \textbf{InstructBLIP}, \textbf{LLaVA-1.5}, and \textbf{Qwen-VL}. As shown in the radar plot (\cref{P-GCCCC}), PND consistently improves performance across all dimensions. Gains in accuracy and fine-grained precision indicate reduced hallucinations (see \cref{P-case}), with further improvements in relevancy and commonsense plausibility. These results suggest stronger visual grounding not only suppresses errors but also yields more coherent, on-topic responses. Full GCCCE results appear in Supplementary~Material~V.

\paragraph{Ablation Study}
To dissect our dual pathways, we conducted an ablation study on the POPE benchmark (\cref{tab4}). We compared our \textbf{PND (Full)} framework against the Baseline and two variants: \textbf{P-only} (amplifying the visual likelihood) and \textbf{N-only} (penalizing the language prior). The results show that both single-pathway variants outperform the baseline, validating their individual effectiveness. More importantly, the full PND model markedly surpasses both, demonstrating that the two pathways are not redundant but strongly synergistic. This synergy supports our hypothesis: the positive pathway acts as an enhancer, enriching visually grounded details, while the negative pathway acts as a suppressor, constraining ungrounded prior-driven hallucinations. Their interplay—simultaneously encouraging evidence-backed detail and discouraging unsupported content—confirms dual-pathway belief adjustment as essential for achieving both fidelity and descriptive richness.

\paragraph{Hyperparameter Analysis}
Supplementary~Material~IV reports a full hyperparameter study. We examine decoding strategies (temperature, top-p, top-k) and key belief-adjustment parameters $\alpha$, $\gamma$, and $\beta$. Results show deterministic or near-deterministic decoding yields strongest visual grounding. Performance is most sensitive to the $\alpha$–$\gamma$ trade-off controlling Bayesian adjustment, while $\beta$ has milder impact. All main experiments use fixed hyperparameters.

\begin{table}[t]
\centering
\caption{Ablation Study on Positive and Negative Components}
\renewcommand{\arraystretch}{0.6} 
\label{tab4}
\begin{tabular}{@{}ccc|cc@{}}
\toprule
\textbf{Original} &\textbf{Positive} & \textbf{Negative} & \textbf{Accuracy} &  \textbf{F1} \\
\midrule
   \checkmark      &          &                   & 78.53                                    & 77.04 \\
\checkmark      &\checkmark        &                   & 83.14                                    & 82.21 \\
 \checkmark      &                 & \checkmark        & 82.23                                   & 80.67 \\
      &\checkmark        & \checkmark        & 83.60     & 82.35 \\
\checkmark       &\checkmark        & \checkmark       & \textbf{84.03}     & \textbf{83.48} \\
\bottomrule
\end{tabular}
\end{table}

\section{Conclusion}
In this work, we addressed object hallucination in MLLMs by identifying it as a
\textbf{Bayesian reasoning imbalance}. We introduced
\textbf{Positive-and-Negative Decoding (PND)}, a training-free framework that
performs real-time Bayesian belief adjustment during generation. PND’s
core mechanism establishes a dynamic contrast: a positive pathway amplifies
the visual likelihood by reinforcing evidence-bearing regions, while a negative
pathway isolates the language prior through a controlled,
evidence-blind counterfactual. This symmetric design effectively suppresses
prior-dominant hallucinations and steers generation toward visually grounded outputs. Extensive experiments on POPE, MME, and CHAIR demonstrate strong hallucination
reduction, while GCCCE results show that PND improves not only grounding but
also factual accuracy and commonsense plausibility. Importantly, although PND
introduces modest inference overhead due to attention extraction, the gains in
visual fidelity and reliability substantially outweigh this cost. Overall, this
work demonstrates that dual-path Bayesian adjustment is a principled and
practical strategy for improving the robustness of modern MLLMs.

\section{Acknowledgement}
This work is sponsored by Beijing Nova Program.
{
    \small
    \bibliographystyle{ieeenat_fullname}
    \bibliography{main}
}

\end{document}